\documentclass{article}

\usepackage[final]{neurips_2023}

\usepackage[utf8]{inputenc} 
\usepackage[T1]{fontenc}    
\usepackage[hidelinks]{hyperref}       
\usepackage{url}            
\usepackage{booktabs}       
\usepackage{amsfonts}       
\usepackage{nicefrac}       
\usepackage{microtype}      
\usepackage{xcolor}         
\usepackage{adjustbox}
\usepackage{multirow}
\usepackage{graphicx}
\usepackage{pgfplots} 
    \usepgfplotslibrary{groupplots}
\usepackage{siunitx}
\usepackage{wrapfig}
\usepackage{booktabs}
\usepackage{pifont}
\usepackage{algorithm}
\usepackage{algpseudocode}
\usepackage{enumitem}
\usetikzlibrary{patterns}
\usepackage{amsmath, amssymb}
\usepackage{cleveref}
\usepackage{subcaption}
\usepackage{caption}
\usepackage{setspace}

\usepackage[compact]{titlesec}
\titlespacing{\subsection}{0pt}{*1}{*0}
\setcitestyle{square}
\setcitestyle{citesep={,}}

\title{Efficiently Adapting Pretrained Language Models to New Languages}

\author{%
  Zoltan Csaki, Pian Pawakapan, Urmish Thakker, Qiantong Xu \\
  SambaNova Systems, Inc. \\
  Palo Alto, CA, USA \\
  \texttt{zoltan.csaki@sambanovasystems.com} \\
}

\begin{document}

\maketitle

\vspace{-1.5em}
\begin{abstract}


Recent large language models (LLM) exhibit sub-optimal performance on low-resource languages, as the training data of these models is usually dominated by English and other high-resource languages. 
Furthermore, it is challenging to train models for low-resource languages, especially from scratch, due to a lack of high quality training data. 
Adapting pretrained LLMs reduces the need for data in the new language while also providing cross lingual transfer capabilities. However, naively adapting to new languages leads to catastrophic forgetting and poor tokenizer efficiency.
In this work, we study how to efficiently adapt any existing pretrained LLM to a new language without running into these issues.
In particular, we improve the encoding efficiency of the tokenizer by adding new tokens from the target language and study the data mixing recipe to mitigate forgetting.
Our experiments on adapting an English LLM to Hungarian and Thai show that our recipe can reach better performance than open source models on the target language, with minimal regressions on English.

\end{abstract}

\section{Introduction \& Related work}
\label{sec:intro}
Multilingual large language models have become prevalent recently \cite{workshop2023bloom, muennighoff2023crosslingual, xue2021mt5, conneau2020unsupervised, goyal2021largerscale, shliazhko2022mgpt}, and have shown strong cross lingual knowledge and capability transfer \cite{lin2022fewshot, xu2021bert, yong2023bloom1, phang2020english, ebrahimi2021adapt, ye2023language, armengolestapé2021multilingual}. However, these multilingual models tend to perform poorly on low-resource languages. On top of this, training models for low-resource languages from scratch is also challenging due to a lack of training data and prohibitive computational requirements. These challenges, along with the prevalence open sourced English models creates an interesting opportunity to see how they can be adapted to new languages quickly, without wasting resources by pretraining from scratch. While prior work \cite{yong2023bloom1, phang2020english, ebrahimi2021adapt, ogueji2021small, armengolestapé2021multilingual} has studied this concept, there are two important questions that warrant further investigation.

\textbf{\emph{How to efficiently encode the new language?}} Byte Pair Encoding (BPE) \cite{gage1994new} tokenizers are commonly used in LLMs including GPT\cite{radford2019language, brown2020language}, Llama \cite{touvron2023llama, touvron2023llama2} and BLOOM \cite{workshop2023bloom, muennighoff2023crosslingual}. These tokenizers are able to encode text at the byte level so that they can generalize to characters that are outside of their vocabulary; this means that any BPE tokenizer can be used for all languages. However, the BPE tokenizer has poor tokenization efficiency if it was not trained on a given language. For example, the original English-centric GPT2 tokenizer with a vocabulary size of 50k needs to use 3.8 times more tokens to encode Thai compared to a smaller tokenizer with a vocabulary size of 5k that is trained on Thai. This will inevitably cost us 3.8 times more compute in both training and inference. Furthermore, it has been shown that models with sub-optimal tokenizers can also have worse evaluation results \cite{rust2021good, stollenwerk2023training}.
In our work, we show how to improve tokenizer fertility\cite{acs2019} by replacing the least frequent tokens in the base model with tokens from the new language.

\textbf{\emph{How to avoid catastrophic forgetting?}} Many works have shown that when continuing to train a LLM on data from a new domain, it undergoes catastrophic forgetting of the original domain it was trained on \cite{french1999catastrophic}, and similar issues appear when training on a new language \cite{french1999catastrophic, yong2023bloom1, cahyawijaya2023instructalign, muennighoff2023crosslingual, chalkidis2021multieurlex, phang2020english, vu2022overcoming}. Different training paradigms including instruction-align\cite{cahyawijaya2023instructalign}, MAD-X \cite{pfeiffer-etal-2020-mad}, (IA)$^3$ \cite{liu2022few} are proposed to alleviate this issue, 
while mixing the training corpus from different languages \cite{yong2023bloom1, ebrahimi2021adapt, ogueji2021small, pires2023sabia, ye2023language, armengolestapé2021multilingual} is an approach shared among all the methods above. 
Thus, in order to avoid forgetting, we study how to use the minimum amount of mixed training data in both continuous pretraining and instruction tuning stages.

We adapt an English-centric model to Hungarian and Thai, and our evaluations show that adding new tokens and mixing training data from both languages can retain the model's English capabilities in addition to improving the models ability to learn the new language. Some contemporary works explore similar, but far less efficient methods of training LLMs on low resource languages. \cite{sengupta2023jais} builds an English-Arabic bilingual LLM, but they train it from scratch; while \cite{pires2023sabia} builds one for English-Portuguese, but it does not optimize the tokenizer or mix the training data.

\section{Implementation Details}
\label{sec:methods}

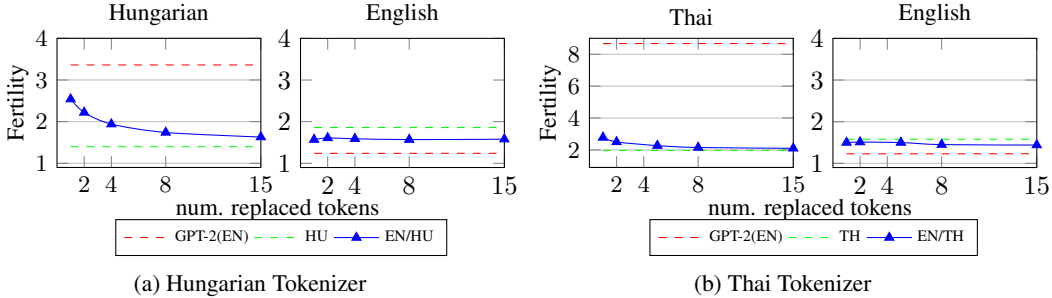
\begin{figure}
\captionof{figure}{Fertility score of the bilingual tokenizers (left: Hungarian, right: Thai) with different number of tokens replaced by new language. The red lines represent the original GPT-2 tokenizer (50k vocabulary), while the green lines represent the tokenizer trained purely on the new language. Every tokenizer has the same total vocabulary size. The number of replaced tokens are in 10$^3$ scale.}\label{fig:fertility}
\vspace{-0.2cm}
\centering
\begin{subfigure}[b]{0.48\textwidth}
\begin{tikzpicture}
\begin{groupplot}[group style = {group size = 2 by 1, horizontal sep = 15pt}, width =\linewidth, height = 3.3cm]
    \nextgroupplot[
        xlabel=$x$,
        ylabel=$y$,
        xmin=0, xmax=15,
        ymin=0.9, ymax=4,
        xtick={2,4,8,15},
        ytick={1,2,3,4},
        width=0.64\linewidth, height=3.3cm,
        title=Hungarian,
        title style={yshift=-1ex, font=\small},
	ylabel=Fertility,
	ymajorgrids,
        y tick label style={font=\small},
        y label style={yshift=-4.5ex, font=\small},
	xlabel=num. replaced tokens,
        x tick label style={font=\small},
        x label style={yshift=1.5ex, xshift=10ex, font=\small},
        legend style={
            at={(1.1,-0.4)},
            anchor=north,
            legend columns=3,
            font=\tiny,
            /tikz/every even column/.append style={column sep=0.cm},
        },
    ]
        \addplot[dashed, smooth, red] plot coordinates {
            (1,3.36)
            (2,3.36)
            (4,3.36)
            (8,3.36)
            (15,3.36)
        };
        \addlegendentry{GPT-2(EN)}
    
        \addplot[dashed,color=green]
            plot coordinates {
            (1,1.398)
            (2,1.398)
            (4,1.398)
            (8,1.398)
            (15,1.398)
        };
        \addlegendentry{HU}
    
        \addplot[smooth,color=blue,mark=triangle*]
            plot coordinates {
            (1,2.54)
            (2,2.22)
            (4,1.94)
            (8,1.74)
            (15,1.63)
        };
        \addlegendentry{EN/HU}
    
    \nextgroupplot[
        xmin=0, xmax=15,
        ymin=0.9, ymax=4,
        xtick={2,4,8,15},
        ytick={1,2,3,4},
        width=0.64\linewidth, height=3.3cm,
        title=English,
        title style={yshift=-1ex, font=\small},
    ]
        \addplot[dashed, smooth,red] plot coordinates {
            (1,1.24)
            (2,1.24)
            (4,1.24)
            (8,1.24)
            (15,1.24)
        };
    
        \addplot[dashed,color=green]
            plot coordinates {
            (1,1.86)
            (2,1.86)
            (4,1.86)
            (8,1.86)
            (15,1.86)
        };
    
        \addplot[smooth,color=blue,mark=triangle*]
            plot coordinates {
            (1,1.57)
            (2,1.61)
            (4,1.59)
            (8,1.57)
            (15,1.58)
        };
\end{groupplot}
\end{tikzpicture}
\vspace{-0.3cm}
\caption{Hungarian Tokenizer}
\end{subfigure}
\hspace{0.2cm}
\begin{subfigure}[b]{0.48\textwidth}
\begin{tikzpicture}
\begin{groupplot}[group style = {group size = 2 by 1, horizontal sep = 15pt}, width =\linewidth, height = 3.3cm]
    \nextgroupplot[
        xlabel=$x$,
        ylabel=$y$,
        xmin=0, xmax=15,
        ymin=0.9, ymax=9,
        xtick={2,4,8,15},
        ytick={2,4,6,8},
        width=0.64\linewidth, height=3.3cm,
        title=Thai,
        title style={yshift=-1ex, font=\small},
	ylabel=Fertility,
	ymajorgrids,
        y tick label style={font=\small},
        y label style={yshift=-4.5ex, font=\small},
	xlabel=num. replaced tokens,
        x tick label style={font=\small},
        x label style={yshift=1.5ex, xshift=10ex, font=\small},
        legend style={
            at={(1.1,-0.4)},
            anchor=north,
            legend columns=3,
            font=\tiny,
            /tikz/every even column/.append style={column sep=0.cm},
        },
    ]
        \addplot[dashed, smooth,red] plot coordinates {
            (1,8.67)
            (2,8.67)
            (5,8.67)
            (8,8.67)
            (15,8.67)
        };
        \addlegendentry{GPT-2(EN)}
    
        \addplot[dashed,color=green]
            plot coordinates {
            (1,1.97)
            (2,1.97)
            (5,1.97)
            (8,1.97)
            (15,1.97)
        };
        \addlegendentry{TH}
    
        \addplot[smooth,color=blue,mark=triangle*]
            plot coordinates {
            (1,2.78)
            (2,2.5)
            (5,2.27)
            (8,2.15)
            (15,2.10)
        };
        \addlegendentry{EN/TH}
    
    \nextgroupplot[
        xmin=0, xmax=15,
        ymin=0.9, ymax=4,
        xtick={2,4,8,15},
        ytick={1,2,3,4},
        width=0.64\linewidth, height=3.3cm,
        title=English,
        title style={yshift=-1ex, font=\small},
	ymajorgrids,
    ]
        \addplot[dashed, smooth,red] plot coordinates {
            (1,1.23)
            (2,1.23)
            (5,1.23)
            (8,1.23)
            (15,1.23)
        };
    
        \addplot[dashed,color=green]
            plot coordinates {
            (1,1.58)
            (2,1.58)
            (5,1.58)
            (8,1.58)
            (15,1.58)
        };
    
        \addplot[smooth,color=blue,mark=triangle*]
            plot coordinates {
            (1,1.50)
            (2,1.51)
            (5,1.50)
            (8,1.45)
            (15,1.44)
        };
\end{groupplot}
\end{tikzpicture}
\vspace{-0.3cm}
\caption{Thai Tokenizer}
\end{subfigure}

\vspace{-1.0cm}
\end{figure}

\subsection{Improving Tokenizer Efficiency} \label{token replacement}
To adapt an existing tokenizer to a new language, tokens from the low resource language can be added to the existing tokenizer's vocabulary to improve its fertility. Fertility is defined as the average number of tokens per word \cite{acs2019}, and details about how we calculated it can be found in appendix \ref{tokenizer fertility}. In our work, instead of extending the tokenizer's vocabulary, we replace the least frequent tokens from it with tokens from the new language. This way, we keep the model capability the same by controlling the vocabulary and embedding table size. In particular, we train a BPE tokenizer on the new language with vocabulary size $k$ and check the number of overlapping tokens $o$ with the original tokenizer. Then we replace the least important $k-o$ non-overlapping tokens from the original tokenizer with the new ones. We also reinitialize the corresponding embeddings in the model. For more details see appendix \ref{token details}.

As shown in figure \ref{fig:fertility}, as the number of replaced tokens $k$ increases, the fertility of the tokenizer approaches the monolingual tokenizer fertility on the new language, with minimal regressions on the English fertility. 
We choose to replace around 5000 tokens, which is only 10\% of the overall vocabulary, because it improves the fertility by 42\% on Hungarian and 73\% on Thai. Note that 50\% fertility drop entails two times faster training and inference.
In addition, replacing more tokens beyond 5000 provides diminishing returns on the fertility, while also increases the difficulty of model adaptation due to having more randomly re-initialized token embeddings.

\subsection{Training Data Mixtures}
Training data for both pretraining and finetuning are prepared following the details in Section \ref{sec:experiments}. Once the datasets are prepared for both languages, we shuffle them at sample level, so that every batch contains text from both languages during training. Note that in our experiments, we do not make any further transformations to either the model or the datasets, after the data is prepared on each side, so that our study is orthogonal and complementary to existing proposed methods \cite{cahyawijaya2023instructalign, pfeiffer-etal-2020-mad, liu2022few, yong2023bloom1, ye2023language} focusing on training paradigm studies.



\section{Experiments}
\label{sec:experiments}
\subsection{Experimental Setup}
Training is done in a two stage pipeline. The first stage is adaptive pretraining (PT) where a base pretrained English 13B GPT-2 model (\ref{base_model}) is continuously trained on a mixture composed of the new language and English. Then, the adapted checkpoint is instruction tuned (IT) on a collection of prompt completion pairs from the new language and English. For more information see appendix \ref{base_model},\ref{sec:dataset_details}

We categorize all evaluation tasks into 4 categories. {\textbf{Multiple Choice}, for this category we append each candidate answer to the prompt and pick the highest probability answer. \textbf{Open-ended Question Answering}, where we let the model generate an answer for each question, and report the average F1 score between the model output and the ground truth. \textbf{Summarization}, where we let the model generate a summary and report the average ROUGE-2 score between the model output and ground truth.
\textbf{Translation}, where we let the model generate translated text and report the BLEU score between the model output and the ground truth. When we report the score for each category, it is the averaged score of all the evaluation tasks that we classified into that category in appendix \ref{sec:evaluation}.


\begin{table}[]
\centering
\caption{Each model is labeled by the language it is adapted for, followed by what style of training was done, The EN PT model is the base model for all following rows. PT stands for pretrained and IT stands for instruction tuned. Each column represents the average of all the benchmarks from a classified under a language and category, the constituent benchmarks can be found in appendix \ref{sec:evaluation}}
\label{tab:main_restult}
\begin{adjustbox}{max width=\textwidth}
\setlength{\tabcolsep}{2pt}
\begin{tabular}{ccccllllll}
\toprule
\multicolumn{1}{l}{} & English                   & \multicolumn{4}{c}{Hungarian}                                                                                            & \multicolumn{4}{c}{Thai}                                                                                                                     \\
\cmidrule(lr){2-2} \cmidrule(lr){3-6} \cmidrule(lr){7-10}
Tasks                & Multi-choice         & Multi-choice         & QA        & \multicolumn{1}{c}{Sum.} & \multicolumn{1}{c}{Trans.} & \multicolumn{1}{c}{Multi-choice} & \multicolumn{1}{c}{QA} & \multicolumn{1}{c}{Sum.} & \multicolumn{1}{c}{Trans.} \\
Metrics                & Acc. ($\uparrow$)         & Acc. ($\uparrow$)         & F1 ($\uparrow$)        & Rouge-2 ($\uparrow$)  & BLEU ($\uparrow$) & Acc. ($\uparrow$) & F1 ($\uparrow$)        & Rouge-2 ($\uparrow$)  & BLEU ($\uparrow$) \\
\midrule
Llama2-7B               &      59.2\%         &     43.9\%         &       4.3\%       &  \multicolumn{1}{c}{2.8}                                   &\multicolumn{1}{c}{}                                 & \multicolumn{1}{c}{47.1\%}                                  & \multicolumn{1}{c}{48.6\%}                                   & \multicolumn{1}{c}{\textbf{30.9}}                                   & \multicolumn{1}{c}{7.7} \\
XGLM-7.5B              &      51.9\%              &    42.8\%          &   \multicolumn{1}{c}{15.8\%}                                   &\multicolumn{1}{c}{0.4}                                 & \multicolumn{1}{c}{1.1}                                  & \multicolumn{1}{c}{46.8\%}                                   & \multicolumn{1}{c}{27.9\%}                                   & \multicolumn{1}{c}{0.0} & \multicolumn{1}{c}{0.2}\\
mt0-xxl              &       52.7\%        &       50.6\%       &    30.6\%          & \multicolumn{1}{c}{2.0}                                   &\multicolumn{1}{c}{\textbf{14.0}}                                 & \multicolumn{1}{c}{46.2\%}                                  & \multicolumn{1}{c}{\textbf{85.3\%}}                                   & \multicolumn{1}{c}{21.9}                                   & \multicolumn{1}{c}{0.9}                                 \\
PULI-GPT-3SX              &     33.5\%          &        43.2\%      &       35.9\%       & \multicolumn{1}{c}{3.1}                                   &\multicolumn{1}{c}{1.3}                                 & \multicolumn{1}{c}{-}                                  & \multicolumn{1}{c}{-}                                   & \multicolumn{1}{c}{-}                                   & \multicolumn{1}{c}{-}                                 \\
openthaigpt-7b-chat              &     \textbf{60.9\%}          &        -      &       -       & \multicolumn{1}{c}{-}                                   &\multicolumn{1}{c}{-}                                 & \multicolumn{1}{c}{43.7\%}                                  & \multicolumn{1}{c}{43.4\%}                                   & \multicolumn{1}{c}{26.5}                                   & \multicolumn{1}{c}{4.0}                                 \\
\midrule
EN PT                & 57.3\%              & 46.2\%              & 32.8\%              & \multicolumn{1}{c}{0.9}                                   &\multicolumn{1}{c}{1.9}                                 & \multicolumn{1}{c}{46.0\%}                                  & \multicolumn{1}{c}{14.4\%}                                   & \multicolumn{1}{c}{0.0}                                   & \multicolumn{1}{c}{0.2}                                 \\
EN PT + IT          & 57.3\%              & 45.9\%              & 34.4\%              & \multicolumn{1}{c}{1.3}         & \multicolumn{1}{c}{1.7}       &  \multicolumn{1}{c}{46.2\%}                                  & \multicolumn{1}{c}{16.1\%}                                   & \multicolumn{1}{c}{2.7}                                   & \multicolumn{1}{c}{0.0}                                \\
HU PT                & 55.3\%              & 44.9\%              & 48.5\%              & \multicolumn{1}{c}{3.7}                                  &  \multicolumn{1}{c}{7.7}                              & \multicolumn{1}{c}{-}                                  & \multicolumn{1}{c}{-}                                   & \multicolumn{1}{c}{-}                                   & \multicolumn{1}{c}{-}                                 \\
HU PT + IT           & 58.0\%              & \textbf{54.9}\%              & \textbf{64.3}\%              & \multicolumn{1}{c}{\textbf{9.2}}         & \multicolumn{1}{c}{6.1}       &  \multicolumn{1}{c}{-}                                & \multicolumn{1}{c}{-}                                    & \multicolumn{1}{c}{-}                                   & \multicolumn{1}{c}{-}                                 \\
TH PT                & 57.4\% & \multicolumn{1}{c}{-}   & \multicolumn{1}{c}{-}  &  \multicolumn{1}{c}{-}                                   & \multicolumn{1}{c}{-}                          & \multicolumn{1}{c}{48.4\%}  & \multicolumn{1}{c}{31.1\%}                                 &                          \multicolumn{1}{c}{11.4}       &                        \multicolumn{1}{c}{3.9}                                \\
TH PT + IT          & 56.4\% & \multicolumn{1}{c}{-}   & \multicolumn{1}{c}{-}   & \multicolumn{1}{c}{-}                                     & \multicolumn{1}{c}{-}                                   &   \multicolumn{1}{c}{\textbf{49.9\%}}                             &     \multicolumn{1}{c}{48.9\%}                              &                      \multicolumn{1}{c}{13.9}             &            \multicolumn{1}{c}{\textbf{12.5}}                     \\
\bottomrule         
\end{tabular}
\end{adjustbox}
\end{table}

\subsection{Main Results}\label{main_results}
We list all the results in table \ref{tab:main_restult}. The HU PT model is trained from EN PT with 50\% HU, 50\% EN data, and TH PT is similarly trained but with Thai data. The HU PT + IT model is trained from the HU PT checkpoint with 50\% HU, 50\% EN IT data. The TH PT + IT is similarly trained from the TH PT checkpoint with 50\% TH, 50\% EN IT data. The EN models trained on purely English data are used as baselines. We list out the dataset and training details in appendix \ref{sec:dataset_details}, \ref{sec:training_details}.

Table \ref{tab:main_restult} shows that with our proposed training recipe, the adapted models are able to maintain the performance on English benchmarks, and improve significantly on the benchmarks of the new languages. This confirms the effectiveness of replacing the tokens in the tokenizer and mixing training data to efficiently adapt a LLM to a new language. On top of this, the adapted models perform as well or better than the state of the art baseline models we have evaluated.

\subsection{Ablation Studies}\label{ablation_study}

\subsubsection{Tokenizer}\label{tokenizer ablation}

\begin{wraptable}[5]{r}{0.36\textwidth}
\vspace{-56pt}
\captionof{table}{Performance of Hungarian model with different tokenizers. 
}\label{tab:tokenizer_ablation}
\begin{adjustbox}{max width=\linewidth}
\begin{tabular}{ccc}
\toprule
\multicolumn{1}{l}{} & \multicolumn{1}{c}{GPT2} & Bilingual \\
\midrule
EN - Multi-choice    &   57.9\%                   & \textbf{58.0\%}   \\
HU- Multi-choice     &  50.8\%                    & \textbf{54.9\%}   \\
HU - QA              &       63.2\%                   & \textbf{64.3\%}   \\
HU - Sum.            &     7.9                    & \textbf{9.2}     \\
HU - Trans.          &  \textbf{8.8}                        & 6.1     \\
\bottomrule
\end{tabular}
\end{adjustbox}
\end{wraptable}

In this experiment, we evaluate models trained on identical data during both the continuous pretraining and IT stages. These models follow the same training recipe but use different tokenizers. Table \ref{tab:tokenizer_ablation} shows that \textbf{the model trained with the bilingual tokenizer performs as well or better than the model that uses the original tokenizer}, on both English and Hungarian tasks. While at the same time, the bilingual tokenizer has much better encoding efficiency as illustrated in Figure \ref{fig:fertility}.

\subsubsection{Pretraining data mixture}\label{pretraining_data_mixture}
Given the same total amount of training data, we tested varying the percentage of English data (50\%, 25\% and 0\%) in the English/Hungarian bilingual data mixture. All training is run for 30k steps. We also compare this to training a pure Hungarian model using only Hungarian data \cite{Nemeskey:2020}, a Hungarian tokenizer, from scratch for 100k steps. All the training details can be found in appendix \ref{pretraining hyperparameters}.

\begin{wrapfigure}[16]{r}{7cm}
\vspace{-12pt}
\caption{Varying pretraining data mixtures. "EN" and "HU" models are monolingual models trained from scratch, while the other models are trained from the "EN" model with the labeled data mixture. }\label{fig:pretraining_mixture}
\begin{tikzpicture}
\begin{groupplot}[group style = {group size = 2 by 1, horizontal sep = 35pt}, width =\linewidth, height = 3.5cm]
    \nextgroupplot[
        ybar=0.0cm,
        bar width=5pt,
        width=0.55\linewidth, height=3.5cm,
        enlarge x limits=0.2,
        title=HU Eval Perplexity,
        title style={yshift=-1ex, font=\small},
	ymajorgrids,
        ylabel={Byte Perplexity $\downarrow$},
        ymin=0,
        xtick=data,
        ytick={0, 1, 2, 3},
        x tick label style={xshift=0.8ex,rotate=45,anchor=east,yshift=-2pt,font=\tiny},
        y label style={yshift=-3.5ex, font=\small},
        symbolic x coords = {EN, 50EN/50HU, 25EN/75HU, 0EN/100HU, HU},
        tickwidth         = 1pt,
        legend style={
            at={(1.1,-0.4)},
            anchor=north,
            legend columns=3,
            font=\tiny,
            /tikz/every even column/.append style={column sep=0.cm},
        },
    ]
        \addplot[black, fill=blue] coordinates { 
            (EN, 2.80)
            (50EN/50HU, 1.84)
            (25EN/75HU, 1.82)
            (0EN/100HU, 1.98)
            (HU, 2.05)
        };
    
    \nextgroupplot[
        ybar=0.0cm,
        bar width=5pt,
        width=0.55\linewidth, height=3.5cm,
        enlarge x limits=0.2,
        title=EN Multi-choice,
        title style={yshift=-1ex, font=\small},
	ymajorgrids,
        ylabel={Accuracy $\uparrow$},
        ymin=0,
        xtick=data,
        ytick={0, 30, 60, 90},
        x tick label style={xshift=0.8ex,rotate=45,anchor=east,yshift=-2pt,font=\tiny},
        y label style={yshift=-3.5ex, font=\small},
        symbolic x coords = {EN, 50EN/50HU, 25EN/75HU, 0EN/100HU, HU},
        tickwidth         = 1pt,
        legend style={
            at={(1.1,-0.4)},
            anchor=north,
            legend columns=3,
            font=\tiny,
            /tikz/every even column/.append style={column sep=0.cm},
        },
    ]
        \addplot[black, fill=blue] coordinates { 
            (EN, 55.9)
            (50EN/50HU, 52.5)
            (25EN/75HU, 54.0)
            (0EN/100HU, 37.2)
            (HU, 43.0)
        };
\end{groupplot}
\end{tikzpicture}
\end{wrapfigure}
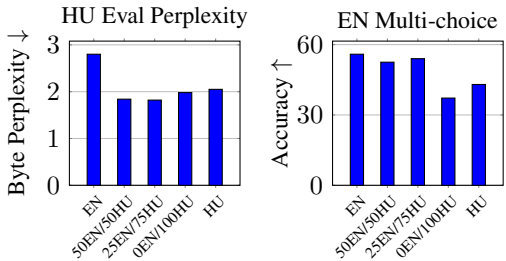

We summarize the comparison results in Figure \ref{fig:pretraining_mixture}. The first finding is that \textbf{it is effective to add in training data from the new language during pretraining}, because those models greatly outperformed the baseline English model. Second, \textbf{it is better to adapt a pretrained LLM than train a new one from scratch}, as the adapted checkpoint performs better on both languages, even though they are trained for one third as long. Third, when comparing the results from the models trained with and without English data mixed in, we can see that \textbf{mixing English data can mitigate the catastrophic forgetting on English and improve the model performance on Hungarian}. Note that there is no significant difference between mixing 50\% and 25\% of English data during training, which implies that adaptation is not sensitive to the exact mixture ratio as long as the original language and new language are included.

\subsubsection{IT data mixture}\label{it_data_mixture}

\begin{wrapfigure}[14]{r}{6.5cm}
\vspace{-16pt}
\caption{Model performance with different IT data mixture. ROUGE-2 score is reported for HU Sum, while accuracy and F1 scores are reported for the rest of the tasks.}\label{fig:IT_mixture}
\vspace{-4pt}
\begin{tikzpicture}
    \begin{axis}[
        xmin=0, xmax=50,
        ymin=30, ymax=70,
        xtick=data,
        symbolic x coords={0, 1, 5, 10, 20, 30, 40, 50},
        axis y line*=left,
        ytick={30,40,50,60,70},
        width=1.03\linewidth, height=3.5cm,
	ylabel=Accuracy/F1,
	ymajorgrids,
        y tick label style={font=\small},
        y label style={yshift=-3ex, font=\small},
	xlabel=Percentage of Hungarian data,
        x tick label style={font=\small},
        x label style={yshift=1.5ex, font=\small},
        legend style={
            at={(0.5,-0.4)},
            anchor=north,
            legend columns=2,
            font=\tiny,
            /tikz/every even column/.append style={column sep=0.cm},
        },
    ]
    \addplot [blue,smooth, error bars/.cd, y dir=both, y explicit] plot coordinates {
        (0, 57.9) +- (0.33, 0.33)
        (1, 57.9) +- (0.41, 0.41)
        (5, 57.8) +- (0.51, 0.51)
        (10,57.7) +- (0.38, 0.38)
        (20,58.0) +- (0.09, 0.09)
        (30,57.2) +- (0.25, 0.25)
        (40,56.7) +- (0.23, 0.23)
        (50,56.5) +- (0.34, 0.34)
    }; \label{plot_one}
    \addlegendentry{EN - multi-choice}

    \addplot[smooth,color=red, error bars/.cd, y dir=both, y explicit]
        plot coordinates {
        (0, 44.3) +- (0.10, 0.10)
        (1, 50.8) +- (1.45, 1.45)
        (5, 50.0) +- (0.39, 0.39)
        (10,49.5) +- (1.11, 1.11)
        (20,51.8) +- (1.54, 1.54)
        (30,54.0) +- (0.02, 0.02)
        (40,51.4) +- (1.40, 1.40)
        (50,52.2) +- (1.62, 1.62)
    }; \label{plot_two}
    \addlegendentry{HU - multi-choice}

    \addplot[smooth,color=green, error bars/.cd, y dir=both, y explicit]
        plot coordinates {
        (0, 40.8) +- (10.20, 10.20)
        (1, 64.3) +- (3.18, 3.18)
        (5, 63.8) +- (4.53, 4.53)
        (10,63.4) +- (4.25, 4.25)
        (20,65.1) +- (2.31, 2.31)
        (30,65.8) +- (2.26, 2.26)
        (40,65.9) +- (2.26, 2.26)
        (50,66.6) +- (1.29, 1.29)
    }; \label{plot_three}
    \addlegendentry{HU - QA}
    \end{axis}

    \begin{axis}[
      width=1.03\linewidth, height=3.5cm,
      axis y line*=right,
      axis x line=none,
      ymin=0, ymax=0.22,
      ylabel=ROUGE-2,
      y tick label style={font=\small},
      y label style={yshift=-45ex, rotate=180, font=\small},
      ytick={0,0.1,0.2},
      xtick=data,
      xmin=0, xmax=50,
      symbolic x coords={0, 1, 5, 10, 20, 30, 40, 50},
      legend style={
          at={(0.5,-0.4)},
          anchor=north,
          legend columns=2,
          font=\tiny,
          /tikz/every even column/.append style={column sep=0.cm},
      },
    ]
    \addlegendimage{/pgfplots/refstyle=plot_one}\addlegendentry{EN - multi-choice}
    \addlegendimage{/pgfplots/refstyle=plot_two}\addlegendentry{HU - multi-choice}
    \addlegendimage{/pgfplots/refstyle=plot_three}\addlegendentry{HU - QA}
    \addplot[smooth,brown, error bars/.cd, y dir=both, y explicit]
      coordinates{
        (0, 0.026) +- (0.0032, 0.0032)
        (1, 0.115) +- (0.0050, 0.0050)
        (5, 0.150) +- (0.0038, 0.0038)
        (10,0.163) +- (0.0038, 0.0038)
        (20,0.170) +- (0.0091, 0.0091)
        (30,0.186) +- (0.0041, 0.0041)
        (40,0.186) +- (0.0031, 0.0031)
        (50,0.180) +- (0.0076, 0.0076)
    }; 
    \addlegendentry{HU - Sum.}
    \end{axis}
\end{tikzpicture}
\end{wrapfigure}
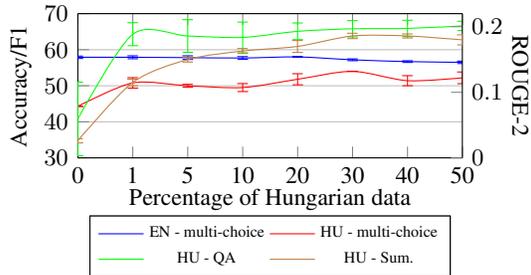 

We run all instruction tuning experiments from the Hungarian pretrained model using 6 gigabytes of instruction tuning text data (2 billion tokens) and the same training settings. Each experiment is repeated 3 times with different random dataset samples. For more details on instruction tuning datasets or training settings see appendix \ref{english instruction tuning data}, \ref{hungarian instruction tuning data} and \ref{instruction tuning hyperparameters}

There is a lack of diverse high quality instruction tuning data in most languages besides English. Thus, we study the impact of the amount of IT data from the new language on the final model performance by varying the Hungarian instruction tuning data mixing rate from  0\% to 50\%. 
Figure \ref{fig:IT_mixture} shows that when no Hungarian instruction tuning data is included, the model undergoes catastrophic forgetting of Hungarian. 
However, the model performance on Hungarian improves as more Hungarian data is mixed in, with marginal returns after more than 1\% of the data is Hungarian. This indicates that a small amount of IT data from the new language gives most of the model performance on the new language.



\section{Conclusion}
\label{sec:conclusion}

In our paper, we study the recipe to efficiently adapt an existing pretrained LLM to a new language, with better tokenizer efficiency and without catastrophic forgetting of its original knowledge.
With only 10\% of tokens in the tokenizer replaced by the the new ones from the target language, it can drop the fertility by 50\% and 70\% on Hungarian and Thai respectively, with limited regression on English. This can greatly improve the efficiency of both training and inference on the new language by 2x and 3x.
In addition, with mixing training data from both languages in pretraining and IT stages, we show that this can improve the model performance on the new language while retaining the model capability on the original language.


\bibliographystyle{IEEEtran}
\bibliography{ref}

\newpage
\appendix

\section{Tokenizer Details}
\label{sec:tokenizer}
\subsection{Tokenizer Fertility Definition}\label{tokenizer fertility}
Fertility is defined as the average number of tokens per word \cite{acs2019}. Words are defined by the Universal Dependencies framework, which gives a "consistent annotation of grammar (parts of speech, morphological features, and syntactic dependencies) across different human languages" \cite{nivre-etal-2016-universal}. To calculate the fertility we tokenize each word from the treebank individually to get the sum total number of tokens, and then divide this by the number of words in the treebank. For Hungarian we used the test set of the "Szeged Dependency Treebank" \cite{vincze-etal-2010-hungarian}, for Thai we used the test set of the Thai "CoNLL 2017" treebank \cite{zeman-etal-2017-conll} and for English we used the test set of the "Gold Standard Universal Dependencies Corpus" \cite{silveira14gold}.

\subsection{Token Replacement Details} \label{token details}

In our work, instead of extending the tokenizer's vocabulary, we replace the least frequent tokens from it with tokens from the new language. This way, we keep the model capacity the same by controlling the vocabulary and embedding table size. 

\begin{enumerate}
\item{}Train a tokenizer limited to a vocabulary size $k$, where $k$ is the number of tokens you want to replace in the original tokenizer
\item{}Find the number of overlapping tokens $o$ between the new tokenizer of vocab size $k$, and the original tokenizer
\item{} Replace the least frequently used $k-o$ tokens from the original tokenizer with the $k-o$ tokens from the new tokenizer. Ensure that all the unchanged tokens from the original tokenizer keep the same vocabulary indices as they had before.
\begin{enumerate}
\item{}note that in the GPT2 tokenizer the tokens in the vocabulary and merges file are ordered from most frequent to least frequent, so we replace the last $k-o$ vocabulary indices in a GPT2 Tokenizer.
\end{enumerate}
\item{} The GPT2 Tokenizer executes the merges rules in the merges.txt file line by line, so to improve the efficiency on the newly added tokens, Add the merges rules from the $k-o$ new tokens to the beginning of the merges.txt file. 
\begin{enumerate}
\item{}Note that various BPE encoding algorithms are implemented without using the merges rules, so ensure that you examine your tokenizer to see how to improves the tokenizer efficiency. 
\end{enumerate}
\item{}Randomize the embeddings of the replaced tokens in the original model so the new embeddings can be learned.
\end{enumerate}

We tested this tokenizer to ensure that the encoding and decoding of text works properly, and figure \ref{fig:fertility} shows that it also improves the fertility.

\section{Base Model}
\label{base_model}
We train our base model with the same tokenizer and architecture as GPT-2 model \cite{radford2019language}. The model has 40 layers of transformer blocks with hidden dimension 5120 and 13 billion parameters in total. The vocabulary size is 50260. The base model was pretrained on 300B English tokens from the PILE\cite{gao2020pile} and C4\cite{raffel2023exploring} datasets, filtered for only natural language English text.

\section{Datasets}
\label{sec:dataset_details}
\subsection{English pretraining data} \label{english pretraining data}
For the continuous pretraining phase, we often mix English data with either Hungarian or Thai. The English data we used is a 100 gigabyte sample of data from the base model pretraining corpus introduced in section \ref{base_model}.

\subsection{English instruction tuning data} \label{english instruction tuning data}
To construct our English instruction tuning dataset, we sample each constituent task from FlanV2 \cite{longpre2023flan} and OIG \cite{Nguyen_2023} equally by raw text size with a fixed dataset size budget. This creates an instruction tuning dataset that is task diverse and reasonably sized. The benefit of sampling instruction tuning data at the task level is shown in  \cite{iyer2023optiml}, and provides a compute efficient alternative to training on all the data. The dataset is 2.6 gigabytes of raw text and about 1.9 million samples.

\subsection{Hungarian pretraining tuning data}\label{hungarian pretraining data}
The dataset used for Hungarian pretraining is the Hungarian Webcorpus 2.0 \cite{Nemeskey:2020}. Our dataset is 96 gigabytes and 11,152,900 documents.

\subsection{Hungarian instruction tuning tuning data}\label{hungarian instruction tuning data}
 There is a lack of naturally written Hungarian instruction tuning datasets, so we use google translate to translate our English Instruction tuning corpus \ref{english instruction tuning data} to Hungarian. During translation the prompt and completion are translated separately and only concatenated during training. 

\subsection{Thai pretraining data}\label{thai pretraining corpus}
For the Thai pre-training corpus, we combine the Thai subsets of OSCAR \cite{abadji2022cleaner}, MC4 \cite{xue-etal-2021-mt5}, and CCNet \cite{wenzek2019ccnet}, which are all derived from Common Crawl. The entire combined corpus was processed with MinHash deduplication \cite{chenghao_mou_2023_8364980, 666900} with 1-grams and a Jaccard similarity of 0.6, with sentence level n-grams (split by whitespace), and totals 15.32 million documents.

\subsection{Thai instruction tuning data}\label{thai instruction tuning corpus}
For Thai instruction tuning data, we use a mixture of manually templated Thai datasets, as well as existing IT datasets translated from English. 

We take various Thai NLP datasets and create a variety of prompting templates for each task to form an instruction tuning dataset. These consist of translation \cite{Nomoto2019InterpersonalMA} \cite{BuschbeckWolf2020APE} \cite{Riza2016IntroductionOT} \cite{Ladhak2020WikiLinguaAN} \cite{cettolo-etal-2012-wit3} \cite{team2022NoLL}, NLI \cite{Conneau2018XNLIEC}, QA \cite{Artetxe:etal:2019} \cite{kobkrit_viriyayudhakorn_2021_4539916}, text categorization, sentiment analysis \cite{bact_2019_3457447}, and summarization \cite{chumpolsathien_2020} \cite{hasan-etal-2021-xl} tasks. This collection of datasets totals 6.26 million instruction tuning examples.

The English-translated IT datasets consist of traditional instruction tuning, multi-turn conversation, and domain-specific QA (i.e. general knowledge, finance, science, mathematics) sourced from collections like FLAN \cite{longpre2023flan}, OIG \cite{Nguyen_2023}, Alpaca \cite{alpaca}, Dolly \cite{DatabricksBlog2023DollyV2}, HC3 \cite{guo-etal-2023-hc3}, and OpenAssistant \cite{Kopf2023OpenAssistantC}, totalling 1.06 million examples.

\section{Training Details}
\label{sec:training_details}

\subsection{Pretraining Hyperparameters}\label{pretraining hyperparameters}
The training process utilized cross-entropy loss to optimize the CLM objective. All training runs shared the same hyperparameters to ensure a fair comparison and to avoid hyperparameter searches. When comparing two pre-training ablations, the runs were trained to token parity, training on the same number of tokens regardless of available data or tokenizer efficiency

The hyperparameters used were batch size = 512, fixed learning rate = 0.000015 and weight decay = 0.1. All the tokens were packed into the training sequences, if they did not fit in a sequence then they would be placed in the next sequence so no training tokens are lost.\footnote{\label{note}\href{https://github.com/sambanova/generative_data_prep}{https://github.com/sambanova/generative\_data\_prep}} An attention mask was applied so that only tokens from the same article attend to each other.

\subsection{Instruction Tuning Hyperparameters}\label{instruction tuning hyperparameters}
All instruction tuning studies share the same hyper-parameters. But ablations comparing runs are not run to step parity, rather they are all trained to 1 epoch to ensure they see all the data.

The hyperparameters used are batch size = 128, fixed learning rate = 0.000015, weight decay = 0.1, grad norm clip = 1.0, and prompt loss weight = 0.0 to ensure that prompts are attended to but not trained on. All the tokens were packed into sequences in a greedy fashion, if they did not fit in a sequence then they would be discarded.\footnotemark[1] An attention mask was applied so that only tokens from the same article attend to each other.

\subsection{Hardware Configuration}\label{hardware}
All training is run on SambaNova's Reconfigurable Data Units (RDU) \cite{9567250}.




\section{Evaluation}
\label{sec:evaluation}
The EAI evaluation harness\cite{eval-harness} is used for all benchmarking. The code\footnote{our open source eval suite link} was adapted for new tasks to evaluate the models on non-English languages. We categorize all the tasks into 4 categories:

\paragraph{Multiple-choice} In this category of tasks, we append each candidate option after the prompt and let the model pick answer with the highest probability. We report average accuracy on each tasks.

\begin{table}[!h]
\centering
\caption{Multiple Choice Evaluation Benchmarks}
\label{tab:table-label}
\begin{tabular}{|c|m{0.8\textwidth}|}
\hline
\textbf{Language} & \textbf{Evaluation Tasks} \\
\hline
English & Lambada \cite{paperno2016lambada}, HellaSwag \cite{zellers2019hellaswag}, Openbookqa \cite{mihaylov2018suit}, Boolq \cite{clark2019boolq}, Arc Easy and Challenge \cite{clark2018think}, PiQA \cite{bisk2019piqa}, ANLI R1 \cite{nie2020adversarial} and Winogrande \cite{sakaguchi2019winogrande}. \\
\hline
Hungarian & HULU evaluation suite \cite{ligetinagy2022hulu}, which is composed of human translated tasks HuCB, HuSST, HuWNLI, HuCOPA, HuCOLA and HuRTE. \\
\hline
Thai &  XCOPA \cite{ponti2020xcopa} and WiseSight Sentiment Analysis \cite{bact_2019_3457447} corpus. Translated versions of HellaSwag \cite{zellers2019hellaswag}, MultiRC \cite{MultiRC2018}, RTE \cite{wang2019superglue}. \\
\hline
\end{tabular}
\end{table}

\paragraph{Open-ended Question Answering} In this category of tasks, we let the model to freely generate completions for each question prompt and we report the average F1 score between the model output and the ground truth answer. 

\begin{table}[!h]
\centering
\caption{Open-ended Question Answering Evaluation Benchmarks}
\label{tab:table-label}
\begin{tabular}{|c|m{0.8\textwidth}|}
\hline
\textbf{Language} & \textbf{Evaluation Tasks} \\
\hline
Hungarian & translated versions of BoolQ \cite{clark2019boolq} and Natural Questions \cite{47761} \\
\hline
Thai &  XQuAD \cite{Artetxe:etal:2019} and a translated version of ReCoRD \cite{Zhang2018ReCoRDBT} \\
\hline
\end{tabular}
\end{table}

\paragraph{Summarization} In this category of tasks, we let the model freely generate a summary for each prompt and we report average ROUGE-2 score between the model output and ground truth. 

\begin{table}[!h]
\centering
\caption{Summarization Evaluation Benchmarks}
\label{tab:table-label}
\begin{tabular}{|c|m{0.8\textwidth}|}
\hline
\textbf{Language} & \textbf{Evaluation Tasks} \\
\hline
Hungarian & Translated version of XSum \cite{narayan2018dont}. \\
\hline
Thai &  ThaiSum \cite{chumpolsathien_2020} \\
\hline
\end{tabular}
\end{table}

\paragraph{Translation} In this category of tasks, we let the model to freely generate translated text and we report BLEU score between the model output and the ground truth answer. 

\begin{table}[!h]
\centering
\caption{Translation Evaluation Benchmarks}
\label{tab:table-label}
\begin{tabular}{|c|m{0.8\textwidth}|}
\hline
\textbf{Language} & \textbf{Evaluation Tasks} \\
\hline
Hungarian & wmt 2009 en-hu and wmt 2009 hu-en \footnote{\href{https://www.statmt.org/wmt09/}{https://www.statmt.org/wmt09/}} datasets \\
\hline
Thai & WIT3 Ted Talks Corpus \cite{cettolo-etal-2012-wit3} \\
\hline
\end{tabular}
\end{table}

\end{document}